\documentclass[11pt]{article}
\usepackage{enumerate}
\usepackage{pdfsync}
\usepackage[OT1]{fontenc}

\usepackage[usenames]{color}
\usepackage{smile}
\usepackage[colorlinks,
            linkcolor=red,
            anchorcolor=blue,
            citecolor=blue
            ]{hyperref}
\usepackage{fullpage}
\usepackage[protrusion=true,expansion=true]{microtype}
\usepackage{pbox}
\usepackage{setspace}
\usepackage{tabularx}
\usepackage{float}
\usepackage{wrapfig,lipsum}
\usepackage{enumitem}
\usepackage{microtype}
\usepackage{graphicx}
\usepackage{subfigure}
\usepackage{pgfplots}
\usepackage{mathtools}
\usepackage{caption}
\usepackage{framed}
\usetikzlibrary{arrows,shapes,snakes,automata,backgrounds,petri}
\usepackage{booktabs} 

\allowdisplaybreaks
\usepackage{colortbl}

\usepackage{xcolor}

\usepackage{makecell}
\usepackage{pifont}

\DeclareMathOperator{\FG}{FG}
\DeclareMathOperator{\LS}{LS}
\DeclareMathOperator{\dc}{dc}
\DeclareMathOperator{\KL}{KL}
\DeclareMathOperator{\Regret}{Regret}

\makeatletter
\newcommand*{\rom}[1]{\expandafter\@slowromancap\romannumeral #1@}
\makeatother
\title{\huge High-Probability Regret Bounds for\\Feel-Good Thompson Sampling}
\author
{
    Xuheng Li\thanks{Department of Computer Science, University of California, Los Angeles, CA 90095, USA; e-mail: {\tt xuheng.li@cs.ucla.edu}}
    ~~
    Qingyue Zhao\thanks{Department of Computer Science, University of California, Los Angeles, CA 90095, USA; e-mail: {\tt zhaoqy24@cs.ucla.edu}}
    ~~
    Quanquan Gu\thanks{Department of Computer Science, University of California, Los Angeles, CA 90095, USA; e-mail: {\tt qgu@cs.ucla.edu}}
}

\ifdefined\final
\usepackage[disable]{todonotes}
\else
\usepackage[textsize=tiny]{todonotes}
\fi
\definecolor{shadecolor}{rgb}{0.78, 0.95, 0.89}

\definecolor{linkcolor}{HTML}{ED1C24}
\definecolor{LightCyan}{rgb}{0.58, 0.94, 0.85}
\hypersetup{colorlinks=true, linkcolor=linkcolor, urlcolor=black}
\usepackage[toc, page, header]{appendix}
\usepackage{titletoc}

\begin{document}
    \date{}
    \maketitle

\begin{abstract}
Feel-good Thompson sampling (FGTS) augments the posterior of vanilla Thompson sampling with an exploration term that favors reward functions with high predicted optimal values. This modification achieves minimax-optimal regret up to logarithmic factors in linear contextual bandits, but existing analyses of its two standard variants establish regret guarantees in expectation. In this paper, we derive the first high-probability regret bounds for both cumulative and current-step FGTS in contextual bandits with general realizable function approximation. Our bounds depend on the corresponding decoupling coefficients and the prior mass near the true reward function. In the $d$-dimensional linear setting, both variants satisfy $\Regret(T)=O\big(d\sqrt T\log(2T)+\sqrt T\log(1/\delta)\big)$ with probability at least $1-\delta$, where $T$ is the time horizon. This bound matches the dimension and horizon dependence of algorithms based on the upper confidence bounds, up to logarithmic factors. Relative to the expected-regret guarantees, the cost of controlling the regret with high probability is confined to additive terms with at most linear dependence on $\log(1/\delta)$. The proof ideas are partially developed by GPT-6 Astra under the authors' guidance.

\end{abstract}

\section{Introduction}

Contextual bandits model sequential decision making in which a learner observes a context, selects an action, and receives feedback only for that action.
Function approximation allows information from observed context--action pairs to inform decisions elsewhere, making learning possible even with large or infinite context and action spaces. Two widely studied approaches to exploration are algorithms based on upper confidence bounds (UCB) and Thompson sampling (TS). UCB-based algorithms choose actions optimistically using confidence sets, while TS samples a reward function and acts greedily according to that sample.

Despite its empirical success~\citep{chapelle2011empirical}, vanilla TS can explore too slowly when the posterior assigns little mass to functions that recognize the value of an unobserved action. Feel-good Thompson sampling (FGTS) addresses this difficulty by adding an exploration term to the posterior that favors functions with high predicted optimal rewards~\citep{zhang2022feel}. This modification achieves $\tilde O(d\sqrt T)$ expected regret in $d$-dimensional linear contextual bandits, matching the minimax rate up to logarithmic factors. There are two standard forms of feel-good exploration: cumulative exploration rewards optimism across past contexts, whereas current-step exploration rewards optimism at the current context~\citep{dann2021provably,li2026variance}. Both combine this preference with a likelihood term that measures agreement with past observations.

Existing regret guarantees for these FGTS variants are established in expectation. An expected-regret bound describes average performance over the learner's randomization and the reward noise, but does not by itself rule out substantially larger regret on individual runs. A high-probability guarantee quantifies this risk through a prescribed failure probability $\delta$. Applying Markov's inequality directly to expected regret gives a bound proportional to $1/\delta$, which can be much larger than the logarithmic confidence penalties in UCB analyses. This motivates the following question:

\begin{center}
    \emph{Is it possible to establish high-probability regret upper bounds for feel-good Thompson sampling in contextual bandits with general function approximation?}
\end{center}

The main challenge is that the confidence-set argument used for UCB-based algorithms does not directly transfer to FGTS. In UCB analyses, one constructs a confidence set containing the true reward function with high probability. Optimistic action selection then bounds each round's regret by the uncertainty within that set, and the analysis controls the accumulated uncertainty~\citep{abbasi2011improved,russo2013eluder}. FGTS instead samples from a posterior whose shape depends on the prior, the observation history, and the feel-good term. Even with linear rewards, the maximum over actions in this term generally destroys a Gaussian posterior form. For a general function class, characterizing the resulting distribution and its random normalizing constant is more difficult still. Moreover, a confidence set containing the truth does not ensure that a posterior draw belongs to that set or is optimistic. Existing high-probability analyses of linear TS exploit specific concentration and optimism properties of the sampling distribution~\citep{agrawal2013thompson,abeille2017linear}; establishing analogous properties for the FGTS posterior is not immediate.

Our approach uses a simple but effective application of Markov's inequality to exponential quantities arising from the posterior. Specifically, if a suitably normalized random quantity $U$ satisfies $\EE[\exp(U)]\le1$, then
\begin{align*}
\PP\big(U>\log(1/\delta)\big)\le\delta.
\end{align*}
The substantive step is to establish such exponential estimates for the losses that remain after decoupling, while accounting for the dependence of posterior samples on past observations. This yields logarithmic confidence penalties without requiring an explicit description of a confidence region for posterior draws. The prior mass near the true reward function controls the normalizing terms that enter these estimates.

The two exploration schemes require different exponential estimates. For cumulative exploration, decoupling first produces conditional posterior averages; a multiplicative bound on successive normalizers controls their sum through a telescoping argument. For current-step exploration, decoupling holds along each realized sequence, and we control an exponential loss evaluated at the sampled function. Here the numerator and denominator of the posterior share the same past reward noise, so their ratio must be handled jointly before taking expectations. Markov's inequality then converts these estimates into high-probability regret bounds for both variants. Section~\ref{sec:highlights} explains the two exponential arguments and how they strengthen the corresponding expected-regret analyses.

We establish general bounds in terms of the corresponding decoupling coefficients and the prior mass in a neighborhood of the true reward function. In the linear setting, under suitable prior coverage and parameter tuning, both variants satisfy, with probability at least $1-\delta$,
\begin{align*}
\Regret(T)=O\big(d\sqrt T\log(2T)+\sqrt T\log(1/\delta)\big).
\end{align*}
Thus FGTS retains the dimension and horizon dependence of its expected-regret guarantees, up to logarithmic factors, with an additive confidence penalty that is at most linear in $\log(1/\delta)$.

\section{Related Work}

\paragraph{Feel-Good Thompson sampling (FGTS).}
\citet{zhang2022feel} introduced FGTS to address insufficient exploration by vanilla Thompson sampling in the frequentist setting. By favoring functions with high predicted optimal rewards, FGTS achieves $\tilde O(d\sqrt T)$ expected regret in linear contextual bandits. Its analysis introduces a decoupling coefficient that reduces the exploration problem to controlling squared prediction errors under independent draws of functions and actions. \citet{fan2023power} developed a unified FGTS framework for variants of linear bandits, including misspecified, nonstationary, and lifelong settings. The feel-good principle has also been extended to reinforcement learning: \citet{dann2021provably} proposed a model-free posterior sampling method with a current-episode optimism term and an online decoupling coefficient, while \citet{agarwal2022model} developed model-based optimistic posterior sampling using likelihood-based model estimation. For contextual dueling bandits, \citet{li2024feel} designed a feel-good term suited to pairwise feedback and established nearly minimax-optimal expected regret in the linear setting. More recently, \citet{li2026variance} distinguished cumulative and current-step feel-good exploration and developed a variance-aware algorithm based on the latter, obtaining expected-regret bounds that adapt to heterogeneous noise levels under general function approximation.

The existing regret analyses of these two FGTS variants focus on expectation over the algorithm's randomization and reward observations. Our work establishes high-probability guarantees for both variants through exponential estimates adapted to their posterior structures, retaining the dimension and horizon dependence of their expected-regret guarantees up to logarithmic factors.

\paragraph{High-probability guarantees for linear contextual bandits.}
Confidence-based exploration has a long history in linear contextual bandits~\citep{auer2002using,chu2011contextual,abbasi2011improved}. \citet{abbasi2011improved} developed self-normalized concentration inequalities for least-squares estimation and used the resulting confidence ellipsoids to analyze OFUL. Its high-probability regret bound scales as $\tilde O(d\sqrt T)$ and accommodates changing, potentially infinite action sets. For a finite set of $K$ actions per round, \citet{chu2011contextual} established an $O\big(\sqrt{dT\log^3(KT\log(T)/\delta)}\big)$ high-probability bound for SupLinUCB. The dependence on the finite action count distinguishes this guarantee from action-count-independent linear bandit bounds. These analyses maintain a confidence event containing the true parameter and bound regret by accumulated confidence widths. Our linear FGTS guarantees match OFUL's dimension and horizon dependence up to logarithmic factors, although the precise dependence on $\delta$ differs. High-probability guarantees also exist for linear TS with appropriately designed sampling distributions~\citep{agrawal2013thompson,abeille2017linear}; these results do not directly characterize the posterior tilted by a feel-good term.

\paragraph{High-probability guarantees under general function approximation.}
For nonlinear reward classes, confidence-based methods require complexity measures that describe how observations constrain predictions at unvisited context--action pairs. \citet{russo2013eluder} introduced the eluder dimension and developed least-squares confidence sets whose accumulated widths can be controlled by this dimension and covering numbers. Their confidence-set containment and width bounds provide a route to high-probability guarantees for optimistic exploration beyond linear models. \citet{foster2018practical} developed regression-oracle implementations of elimination and optimistic exploration, with guarantees under additional distributional conditions. A complementary approach is SquareCB~\citep{foster2020beyond}, which obtains high-probability regret bounds for realizable contextual bandits with finitely many actions by reducing the problem to online regression and explicitly randomizing action selection. SquareCB does not rely on UCB action selection or posterior sampling. Our results address the posterior-sampling approach under general function approximation: decoupling controls exploration complexity, prior coverage controls statistical complexity, and exponential estimates control the regret with high probability.

\section{Preliminaries}

\subsection{Contextual Bandits}\label{sec:contextual_bandits}

We consider the contextual bandits setting~\citep{langford2007epoch}. Let $\cX$ be the context set, $\cA$ be the action set, and $T\ge1$ be the total number of rounds. We write $[T]=\{1,\ldots,T\}$. In round $t\in[T]$, the learner receives a context $x_t\in\cX$ and a feasible action set $\cA_t\subseteq\cA$. The learner then selects an action $a_t\in\cA_t$ and receives the randomized reward
\begin{align*}
r_t=f_*(x_t, a_t)+\epsilon_t,
\end{align*}
where $f_*:\cX\times\cA\to[-1, 1]$ is the ground truth reward function, and $\epsilon_t$ is the zero-mean noise.
The goal is to minimize the regret
\begin{align*}
\Regret(T)=\sum_{t=1}^T[f_*(x_t, a_t^*)-f_*(x_t, a_t)],
\end{align*}
where $a_t^*\in\argmax_{a\in\cA_t}f_*(x_t, a)$.

\noindent\textbf{Subgaussian noise.}
Let $\cF_0$ be the initial information and define
\begin{align*}
\cG_t&=\sigma(\cF_{t-1},x_t,\cA_t,f_t,a_t),\qquad
\cF_t=\sigma(\cG_t,r_t).
\end{align*}
Here $f_t$ is the function sampled by the learner, as specified below. Thus $\cG_t$ is the information available after action selection and before observing $r_t$. Following previous works on feel-good Thompson sampling (FGTS), we use the following normalization of subgaussian noise.
\begin{assumption}\label{ass:subgaussian}
For any $t\in[T]$ and $\lambda\in\RR$, the noise $\epsilon_t$ satisfies
\begin{align*}
\log\EE[\exp(\lambda\epsilon_t)|\cG_t]\le\lambda^2/8\quad\text{almost surely}.
\end{align*}
\end{assumption}

\noindent\textbf{Reward function class.}
The learner estimates the reward function using a class $\cF\subseteq\{f:\cX\times\cA\to[-1,1]\}$.
We assume that the function class is realizable, i.e., $f_*\in\cF$.
We consider the linear function class as an example, defined as $\cF_{\mathrm{lin}}=\{f_{\btheta}:\btheta\in\Theta\}$ where $\Theta\subset B(0, 1)$, $f_{\btheta}(x, a)=\langle\btheta, \bphi(x, a)\rangle$. Here $B(0,1)=\{\bv\in\RR^d:\|\bv\|_2\le1\}$, $d\ge1$, $\btheta$ is the model parameter, and $\bphi:\cX\times\cA\to B(0, 1)$ is a feature mapping.

\subsection{Feel-Good Thompson Sampling}

In vanilla Thompson sampling, the agent draws samples of the estimated reward function $f_t\in\cF$ from the posterior distribution $p_t$:
\begin{align*}
p_t(f)\propto p_0(f)\cdot\exp\bigg(-\eta\sum_{s=1}^{t-1}(r_s-f(x_s, a_s))^2\bigg),
\end{align*}
where $p_0$ is a fixed prior distribution, and $\eta>0$ is a tunable hyperparameter. After drawing $f_t$, the learner chooses
\begin{align*}
a_t\in\argmax_{a\in\cA_t}f_t(x_t,a).
\end{align*}
We assume that the displayed maxima are attained and use fixed measurable tie-breaking rules. The posterior formulas describe the conditional law of $f_t$ given $\cF_{t-1},x_t,\cA_t$; densities are taken with respect to a common dominating measure. However, vanilla Thompson sampling can suffer suboptimal regret even for linear reward functions~\citep{zhang2022feel}. Feel-good Thompson sampling addresses this limitation by adding an exploration term to the exponent of the posterior density, of the form
\begin{align*}
\lambda\max_{a\in\cA_t}f(x_t, a),
\end{align*}
where $\lambda>0$ is a tunable hyperparameter.

The intuition is to favor plausible reward functions that predict a high achievable reward. The squared-loss term evaluates a function only on the actions that have been selected. Consequently, functions that agree on frequently selected actions but disagree on an unexplored action can remain indistinguishable from the observed rewards. If most posterior mass underestimates that unexplored action, vanilla Thompson sampling may select it too rarely to correct this pessimism. Feel-good exploration increases the relative weight of functions under which an action offers a higher reward, encouraging the learner to test these optimistic predictions~\citep{zhang2022feel}. When such a function is sampled, the greedy action rule directs the learner toward its predicted best action. Subsequent observations can then penalize inaccurate predictions through the squared-loss term. Thus the posterior balances agreement with the data against the prospect of a larger reward, with $\lambda$ controlling the strength of the latter preference. This bias promotes exploration while retaining random sampling over reward functions; it does not require every sampled function to be optimistic.

\citet{li2026variance} identified two types of feel-good exploration terms:
\begin{itemize}[leftmargin=*]
\setlength{\emergencystretch}{3em}
\item \textbf{Type A:} Cumulative feel-good exploration~\citep{zhang2022feel, agarwal2022model}, which is the sum of $\max_{a\in\cA_s}f(x_s, a)$ in all previous steps:
\begin{align*}
p_t^A(f)\propto p_0(f)\cdot\exp\bigg(\sum_{s=1}^{t-1}\Big[-\eta(r_s-f(x_s, a_s))^2+\textcolor{red}{\lambda\max_{a\in\cA_s}f(x_s, a)}\Big]\bigg).
\end{align*}
\item \textbf{Type B:} Current-step feel-good exploration~\citep{dann2021provably}, which only contains $\max_{a\in\cA_t}f(x_t, a)$ for the current step:
\begin{align*}
p_t^B(f)\propto p_0(f)\cdot\exp\bigg(-\eta\sum_{s=1}^{t-1}(r_s-f(x_s, a_s))^2+\textcolor{red}{\lambda\max_{a\in\cA_t}f(x_t, a)}\bigg).
\end{align*}
\end{itemize}

The two variants differ in where they encourage optimism. Type A rewards functions that predict high achievable rewards across past contexts, accumulating this preference as observations arrive. Type B evaluates the exploration bonus at the current context and feasible action set, so its optimistic bias responds directly to the decision at hand. Both variants use all past reward observations to assess fit, but they combine this evidence with different measures of the potential benefit of exploration. This difference in the placement of the feel-good term leads to the distinct decoupling and concentration arguments developed below.

\section{Main Results}

\subsection{Decoupling Coefficient and Prior Coverage}\label{sec:complexity}

The analyses of Type-A (cumulative) and Type-B (current-step) FGTS use the following decoupling coefficients to characterize the complexity of the function class.

\begin{definition}\label{def:dc_A}
The decoupling coefficient for FGTS with the cumulative feel-good exploration term is the smallest nonnegative number $\dc$ such that, for every $\gamma>0$, context $x\in\cX$, and joint distribution $P$ over $\cF\times\cA$, we have
\begin{align*}
\EE_{(f, a)\sim P}[f(x, a)-f_*(x, a)]\le\frac{\dc}{4\gamma}+\gamma\cdot\EE_{a\sim P}\EE_{f\sim P}[(f(x, a)-f_*(x, a))^2].
\end{align*}
The expectations on the right use independent draws from the two marginals of $P$.
\end{definition}
\begin{definition}\label{def:dc_B}
Let $Z=(z_1,\ldots,z_T)$, where $z_t=(x_t,a_t)\in\cX\times\cA$. The decoupling coefficient for FGTS with the current-step feel-good exploration term is $\dc_T(\cF)=\sup_Z\dc(\cF,Z)$, where $\dc(\cF,Z)$ is the smallest nonnegative number such that, for every $0<\gamma\le1$ and every sequence $f_1,\ldots,f_T\in\cF$, we have
\begin{align*}
\sum_{t=1}^T(f_t(z_t)-f_*(z_t))\le\gamma\sum_{t=1}^T\sum_{s=1}^{t-1}(f_t(z_s)-f_*(z_s))^2+\frac{\dc(\cF, Z)}{4\gamma}.
\end{align*}
This is the contextual-bandit form of the online decoupling inequality in \citet{dann2021provably}. We abbreviate $\dc_T(\cF)$ as $\dc_T$.
\end{definition}

\paragraph{Decoupling in the linear setting.}
For the linear class $\cF_{\mathrm{lin}}$ in Section~\ref{sec:contextual_bandits}, the two coefficients satisfy
\begin{align*}
\dc\le d,\qquad \dc_T=O(d\log(2T)).
\end{align*}
The first bound follows from the covariance argument underlying the linear decoupling bound of \citet{fan2023power,zhang2022feel}. To see this for the joint distributions in Definition~\ref{def:dc_A}, fix $x$ and let $\bSigma=\EE_{a\sim P}[\bphi(x,a)\bphi(x,a)^\top]$. Cauchy--Schwarz in the geometry of $\bSigma$, using its pseudoinverse if it is singular, gives
\begin{align*}
\EE_{(f,a)\sim P}[f(x,a)-f_*(x,a)]
&\le\sqrt{d\cdot\EE_{a\sim P}\EE_{f\sim P}[(f(x,a)-f_*(x,a))^2]}\\
&\le\frac{d}{4\gamma}+\gamma\EE_{a\sim P}\EE_{f\sim P}[(f(x,a)-f_*(x,a))^2].
\end{align*}
Here the dimension factor comes from $\EE_a[\bphi(x,a)^\top\bSigma^\dagger\bphi(x,a)]=\operatorname{rank}(\bSigma)\le d$.
The second bound follows from the elliptical-potential argument used for the online decoupling coefficient in \citet{dann2021provably}. In the present normalization, one can use the regularized design matrix $\bV_t=T^{-1}\bI+\sum_{s<t}\bphi(z_s)\bphi(z_s)^\top$: the accumulated clipped leverage scores $\min\{1,\bphi(z_t)^\top\bV_t^{-1}\bphi(z_t)\}$ are $O(d\log(2T))$. Together with $\|\btheta_t-\btheta_*\|_2\le2$ and $\gamma\le1$, this yields the stated bound uniformly over the sequences in Definition~\ref{def:dc_B}. Thus both coefficients scale linearly with dimension, up to a logarithmic factor, without an explicit dependence on the number of actions.

\paragraph{Expectation versus sequential control.}
Definition~\ref{def:dc_A} addresses the dependence between a sampled function and the action selected from it. Its left-hand side averages their coupled prediction error, while its right-hand side evaluates squared error under independent draws from the same marginals. This is a statement about expectations at a single context. In the regret analysis, it is applied conditionally on the history, so an additional concentration argument is needed to pass from conditional expected regret to realized regret. Definition~\ref{def:dc_B}, in contrast, holds for every realized sequence of functions and context--action pairs, without averaging over posterior samples. It compares the current prediction error of $f_t$ with the errors of that same function on all previously visited pairs. Intuitively, a function that fits the past can make a large new error only along directions that have not been sufficiently explored; the design matrix tracks how these directions are resolved over time.

These structures align with the two exploration terms. Type A accumulates optimism across rounds, and its posterior update supports a telescoping argument for conditional posterior averages. Type B favors optimism at the current context, while its likelihood tests the current sampled function against the entire observation history. Its decoupling inequality therefore controls a sequence of sampled functions directly. The logarithmic horizon dependence in the linear Type-B bound reflects this sequential accumulation of information; Type A instead pays a dimension-dependent cost for decoupling one joint distribution at a time.

\paragraph{Prior coverage.} Additionally, for any $h>0$, let $\cF_h$ be a neighborhood of $f_*$ defined as
\begin{align*}
\cF_h=\{f\in\cF:\|f-f_*\|_\infty\le h\},\qquad
\|f-f_*\|_\infty=\sup_{(x,a)\in\cX\times\cA}|f(x,a)-f_*(x,a)|.
\end{align*}
We write $p_0^{-1}(\cF_h)=1/p_0(\cF_h)$ and use the quantity $\log p_0^{-1}(\cF_h)$ in the regret bounds, which characterizes the coverage of the neighborhood $\cF_h$ by the prior distribution.

\paragraph{Prior coverage in the linear setting.}
This quantity also has a dimension-dependent bound under a suitable prior, as in the parametric prior-mass analyses of \citet{zhang2022feel,dann2021provably}. For a concrete choice, take $\Theta=B(0,1)$ and let $p_0$ be induced by the uniform distribution on this ball. Since $\|\bphi(x,a)\|_2\le1$, we have $\|f_{\btheta}-f_{\btheta_*}\|_\infty\le\|\btheta-\btheta_*\|_2$. For $0<h\le1$, the ball
\begin{align*}
B\big((1-h/2)\btheta_*,h/2\big)\subseteq B(0,1)\cap B(\btheta_*,h)
\end{align*}
therefore consists entirely of parameters whose functions belong to $\cF_h$. Comparing volumes gives
\begin{align*}
p_0(\cF_h)\ge(h/2)^d,\qquad
\log p_0^{-1}(\cF_h)\le d\log(2/h).
\end{align*}
In particular, at $h=1/T$ we obtain $\log p_0^{-1}(\cF_{1/T})\le D_T \coloneqq d\log(2T)=\tilde O(d)$. More generally, the same order holds whenever the parameter prior assigns at least $(ch)^d$ mass to $B(\btheta_*,h)\cap\Theta$, for a fixed $c>0$.

\subsection{Cumulative Feel-Good Exploration}

\begin{theorem}\label{thm:A}
Under Assumption~\ref{ass:subgaussian}, suppose that $0<\dc<\infty$ and $0<p_0(\cF_{1/T})<1$. For $\delta\in(0,1/3)$, choose the hyperparameters of FGTS with the cumulative feel-good exploration term as
\begin{align*}
\eta=0.2,\quad\lambda=\sqrt{\frac{\log p_0^{-1}(\cF_{1/T})}{5T\cdot\dc}}.
\end{align*}
Then with probability at least $1-3\delta$, the regret satisfies
\begin{align*}
\Regret(T)
&\le2\sqrt{5T\cdot\dc\cdot\log p_0^{-1}(\cF_{1/T})}
+4\sqrt{\frac{T\log p_0^{-1}(\cF_{1/T})}{5\dc}}\\
&\quad+\bigg(4\sqrt T+\sqrt{\frac{\dc}{5\log p_0^{-1}(\cF_{1/T})}}\bigg)\sqrt{2\log(1/\delta)}\\
&\quad+\sqrt{\frac{5T\dc}{\log p_0^{-1}(\cF_{1/T})}}\log(1/\delta)
+\sqrt{\frac{\dc}{5T\log p_0^{-1}(\cF_{1/T})}}+1.
\end{align*}
\end{theorem}

\paragraph{Implications for linear bandits.}
Consider the uniform-prior linear setting of Section~\ref{sec:complexity}. The proof of Theorem~\ref{thm:A} allows us to use the deterministic upper bounds $d$ and $D_T=d\log(2T)$ for the decoupling coefficient and log inverse prior mass before optimizing $\lambda$. Taking $\eta=0.2$ and $\lambda=\sqrt{D_T/(5Td)}$, and applying the proof with failure parameter $\delta/3$, gives, for any $\delta\in(0,1)$, with probability at least $1-\delta$,
\begin{align*}
\Regret(T)
&=O\bigg(d\sqrt{T\log(2T)}+\sqrt{T\log(3/\delta)}
+\sqrt{\frac{T}{\log(2T)}}\log(3/\delta)\bigg)\\
&=\tilde O(d\sqrt T),
\end{align*}
where $\tilde O$ suppresses logarithmic factors in $T$ and $1/\delta$. Using upper bounds for tuning avoids requiring a lower bound on $\log p_0^{-1}(\cF_{1/T})$. The leading term combines a dimension factor from decoupling with another from prior coverage. The result retains the dimension and horizon dependence of the expected-regret guarantee for linear FGTS in \citet{zhang2022feel}, while controlling the realized regret with only logarithmic dependence on the inverse failure probability.

\subsection{Current-Step Feel-Good Exploration}

\begin{theorem}\label{thm:B}
Under Assumption~\ref{ass:subgaussian}, suppose that $\dc_T>0$ and $0<p_0(\cF_{1/T})<1$. For $\delta\in(0,1)$, choose the hyperparameters of FGTS with the current-step feel-good exploration term as
\begin{align*}
\eta=0.5,\quad\lambda=\max\Big\{\frac14, \sqrt{T\dc_T^{-1}\log(p_0^{-1}(\cF_{1/T}))}\Big\}.
\end{align*}
Then with probability at least $1-\delta$ the regret satisfies
\begin{align*}
\Regret(T)\le2\sqrt{T\dc_T\log(p_0^{-1}(\cF_{1/T}))}+3+\sqrt{\frac{T\dc_T}{\log(p_0^{-1}(\cF_{1/T}))}}\log(T/\delta).
\end{align*}
\end{theorem}

\paragraph{Implications for linear bandits.}
For the same linear class and uniform prior, let $\overline\dc_T=C d\log(2T)$ be the upper bound from Section~\ref{sec:complexity}, where $C$ is a sufficiently large universal constant, and let $D_T=d\log(2T)$. Using these bounds in the proof of Theorem~\ref{thm:B}, with $\eta=0.5$ and $\lambda=\max\{1/4,\sqrt{TD_T/\overline\dc_T}\}$, gives, with probability at least $1-\delta$,
\begin{align*}
\Regret(T)
&=O\big(d\sqrt T\log(2T)+\sqrt T\log(T/\delta)\big)
=\tilde O(d\sqrt T).
\end{align*}
Thus current-step exploration also achieves the same dimension and horizon dependence as the expected-regret linear specialization of \citet{dann2021provably}, now with a high-probability guarantee. Relative to Type A, the additional logarithmic factor in the decoupling coefficient affects only logarithmic factors in the regret. Although the Type-B decoupling step is deterministic, the posterior samples and reward observations remain random, so concentration is still needed to control the remaining loss terms. Both exploration schemes consequently attain $\tilde O(d\sqrt T)$ regret with high probability through different forms of decoupling.

\section{Highlights of Proof}\label{sec:highlights}

An expected-regret bound gives a tail bound through Markov's inequality at the cost of a factor $1/\delta$. To obtain logarithmic dependence on $1/\delta$, we establish exponential bounds for the losses left by decoupling. The quantities requiring control differ between the two algorithms: Type A produces a random sum of conditional posterior averages, whereas Type B produces losses evaluated at the sampled functions. We write
\begin{align*}
\LS_t(f)=(f(x_t,a_t)-f_*(x_t,a_t))^2,\qquad
\FG_t(f)=\max_{a\in\cA_t}f(x_t,a)-f_*(x_t,a_t^*).
\end{align*}
The greedy action rule gives the regret identity $f_*(x_t,a_t^*)-f_*(x_t,a_t)=f_t(x_t,a_t)-f_*(x_t,a_t)-\FG_t(f_t)$.

\subsection{Type A: exponential control of conditional losses}

Although Definition~\ref{def:dc_A} is stated in expectation, it holds for every joint law of a function and an action. We can therefore apply it conditionally on $\cF_{t-1},x_t,\cA_t$. For $\gamma>0$, let
\begin{align*}
b_t=\EE\Big[\EE_{\tilde f\sim p_t^A}[\gamma\LS_t(\tilde f)-\FG_t(\tilde f)]\Big|\cF_{t-1}\Big],
\end{align*}
where $\tilde f$ is drawn from the posterior independently of $a_t$ under this conditioning. Combining the conditional decoupling inequality with Azuma--Hoeffding's inequality gives, with probability at least $1-\delta$,
\begin{align}
\Regret(T)\le\frac{T\cdot\dc}{4\gamma}+\sum_{t=1}^T b_t+\sqrt{32T\log(1/\delta)}.\label{eq:overview_A_regret}
\end{align}
The sum $\sum_t b_t$ is still random. The expected-regret analysis reviewed in \citet[Section~6.1]{li2026variance} controls the loss by telescoping expected log normalizers. To obtain a tail bound for this sum, we instead prove a conditional multiplicative bound for the normalizer itself.

Let $Z_t^A$ be the normalizing integral of the Type-A posterior after subtracting its exponent at $f_*$, as defined in \eqref{eq:def_ZtA}. Lemma~\ref{lemma:telescope} establishes the following estimate for $0<\eta\le0.2$:
\begin{align*}
\frac{\EE[Z_{t+1}^A|\cF_{t-1}]}{Z_t^A}
&\le\exp\Big(\EE\Big[\EE_{f\sim p_t^A}[-\eta\LS_t(f)/4+\lambda\FG_t(f)]\Big|\cF_{t-1}\Big]+4\lambda^2\Big)\\*
&\le e^{-\lambda b_t+4\lambda^2},
\end{align*}
where the second inequality uses $\gamma\lambda\le\eta/4$. For the first inequality, we average the current reward noise using Assumption~\ref{ass:subgaussian}, then apply the bounded-loss argument from the expected-regret analysis. This bounds the squared-loss exponential in terms of its mean; the bounded feel-good term contributes $4\lambda^2$. Retaining the conditioning on the history places exactly the random $b_t$ from \eqref{eq:overview_A_regret} in the exponent.

The factor $e^{-\lambda b_t}$ cancels round $t$'s contribution to $e^{\lambda\sum_{s=1}^t b_s}$. Since $b_t$ is $\cF_{t-1}$-measurable, multiplying the one-step estimate by $Z_t^A\exp(\lambda\sum_{s=1}^t b_s-4t\lambda^2)$ and iterating from $Z_1^A=1$ gives
\begin{align*}
\EE\bigg[Z_{T+1}^A\exp\bigg(\lambda\sum_{t=1}^T b_t-4T\lambda^2\bigg)\bigg]\le1.
\end{align*}
Markov's inequality therefore yields, with probability at least $1-\delta$,
\begin{align*}
\sum_{t=1}^T b_t\le\frac{-\log Z_{T+1}^A+\log(1/\delta)}{\lambda}+4T\lambda.
\end{align*}
Lemma~\ref{lemma:ZTA} bounds $-\log Z_{T+1}^A$ using the prior mass of $\cF_h$. Combining that estimate with the preceding Markov bound and \eqref{eq:overview_A_regret}, with $h=1/T$ and $\gamma=\eta/(4\lambda)$, gives Theorem~\ref{thm:A}.

\subsection{Type B: controlling the noise in the posterior ratio}

For Type B, Definition~\ref{def:dc_B} gives the pathwise reduction
\begin{align*}
\Regret(T)\le\frac{\dc_T}{4\gamma}+\sum_{t=1}^T Q_t(f_t),\qquad
Q_t(f)=\gamma\sum_{s=1}^{t-1}\LS_s(f)-\FG_t(f),
\end{align*}
where $0<\gamma\le1$. The expected-regret argument in \citet[Section~6.2]{li2026variance} uses fixed-function exponential estimates and KL-regularized optimality to bound a loss averaged over the posterior and the history. Here we bound $\EE e^{\lambda Q_t(f_t)}$, keeping the sampled function inside the exponential. Since $f_t$ depends on past rewards, a fixed-function noise bound cannot be applied directly at $f_t$.

Substituting the posterior density in \eqref{eq:def_ptB} into $\EE e^{\lambda Q_t(f_t)}$ cancels the current feel-good factor and leaves integration over fixed functions under $p_0$. The remaining numerator and denominator $Z_t^B$ contain the same past noises. To use the prior mass near $f_*$, we bound their ratio before taking expectation. Fix $h>0$ with $p_0(\cF_h)>0$, let $p_0^h$ be the prior conditioned on $\cF_h$, and write $\bar f_h(x,a)=\EE_{f\sim p_0^h}f(x,a)$.

Restricting the normalizer to $\cF_h$ and applying Jensen's inequality gives a pathwise bound on $1/Z_t^B$ (Lemma~\ref{lemma:B_Zt}). The exponent in this bound contains the noise term $-2\eta\sum_{s<t}\epsilon_s(\bar f_h(x_s,a_s)-f_*(x_s,a_s))$. Combining it with the numerator changes the noise coefficient from $f-f_*$ to $f-\bar f_h$. With $z_s=(x_s,a_s)$, the resulting estimate is
\begin{align*}
\EE e^{\lambda Q_t(f_t)}
\le\frac{e^{\lambda h}}{p_0(\cF_h)}\EE_{f\sim p_0}\EE\exp\bigg(\sum_{s<t}\big[2\eta\epsilon_s(f(z_s)-\bar f_h(z_s))+(\gamma\lambda-\eta)\LS_s(f)+\eta h^2\big]\bigg).
\end{align*}
Since $p_0^h$ is supported on $\cF_h$, we have $|\bar f_h(z_s)-f_*(z_s)|\le h$. For fixed $f$, the new noise coefficient is $\cG_s$-measurable and, by this bound, satisfies $(f(z_s)-\bar f_h(z_s))^2\le2\LS_s(f)+2h^2$. By Assumption~\ref{ass:subgaussian},
\begin{align*}
&\EE\Big[e^{2\eta\epsilon_s(f(z_s)-\bar f_h(z_s))+(\gamma\lambda-\eta)\LS_s(f)+\eta h^2}\Big|\cG_s\Big]\\*
&\le\exp\big((\gamma\lambda-\eta+\eta^2)\LS_s(f)+(\eta+\eta^2)h^2\big).
\end{align*}
The negative squared-loss term therefore absorbs the noise contribution when $\gamma\lambda\le\eta-\eta^2$. Iterating this estimate and integrating over the prior gives, for $0<\eta\le0.5$,
\begin{align*}
\EE e^{\lambda Q_t(f_t)}\le\frac{e^{\lambda h+0.75(t-1)h^2}}{p_0(\cF_h)}.
\end{align*}
Markov's inequality with failure probability $\delta/T$ and a union bound imply, simultaneously for $t\in[T]$,
\begin{align*}
Q_t(f_t)\le h+\frac{0.75(t-1)h^2+\log p_0^{-1}(\cF_h)+\log(T/\delta)}{\lambda}
\end{align*}
with probability at least $1-\delta$. Summation in the regret inequality, with $h=1/T$ and $\gamma=1/(4\lambda)$, gives Theorem~\ref{thm:B}.

\section{Conclusion}

We established high-probability regret bounds for feel-good Thompson sampling in contextual bandits with general realizable function approximation. Our results cover both cumulative and current-step feel-good exploration, with bounds governed by the corresponding decoupling coefficient and the prior mass near the true reward function. In the linear setting, suitable prior coverage and parameter tuning yield $\tilde O(d\sqrt T)$ regret with high probability for both variants, preserving the dimension and horizon dependence of their expected-regret guarantees while introducing only logarithmic dependence on the inverse failure probability.

The analyses distinguish the probabilistic structures of the two exploration schemes. For cumulative exploration, we control conditional posterior averages through an exponential estimate that telescopes across rounds. For current-step exploration, we combine a pathwise decoupling inequality with exponential control of losses at the sampled functions, accounting for their dependence on past reward noise. These arguments show how the structural complexity measures used in expected-regret analyses can also support high-probability guarantees. Extending the guarantees to reinforcement learning and the variance-dependent setting is a natural direction for future work.

\section{Acknowledgement}
 The proof ideas are partially developed by GPT-6 Astra under the authors' guidance under stronger assumptions. The authors simplified the arguments and re-organized the proof, verified the proof details, prepared the draft, and take full responsibility for its contents.

\appendix

\section{Proof of Main Results}\label{app:proof_main}

We use $\LS_t$, $\FG_t$, and $p_0^h$ from Section~\ref{sec:highlights}, and the filtrations from Section~\ref{sec:contextual_bandits}.
We can then rewrite $p_t^A$ and $p_t^B$ with $\LS_t$ and $\FG_t$. Since $r_s=f_*(x_s,a_s)+\epsilon_s$, we have
\begin{align*}
-\eta(r_s-f(x_s,a_s))^2+\eta(r_s-f_*(x_s,a_s))^2
=2\eta\epsilon_s(f(x_s,a_s)-f_*(x_s,a_s))-\eta\LS_s(f).
\end{align*}
Subtracting the exponent evaluated at $f_*$ from each posterior exponent changes only its normalizing constant.
Therefore,
\begin{align}
p_t^A(f)&=\frac{p_0(f)}{Z_t^A}\exp\bigg(\sum_{s=1}^{t-1}\big[2\eta\epsilon_s(f(x_s, a_s)-f_*(x_s, a_s))-\eta\LS_s(f)+\lambda\FG_s(f)\big]\bigg).\label{eq:def_PtA}
\end{align}
The normalizing integral is
\begin{align}
Z_t^A=\EE_{f\sim p_0}\exp\bigg(\sum_{s=1}^{t-1}\big[2\eta\epsilon_s(f(x_s,a_s)-f_*(x_s,a_s))-\eta\LS_s(f)+\lambda\FG_s(f)\big]\bigg).\label{eq:def_ZtA}
\end{align}
Similarly, we have
\begin{align}
p_t^B(f)&=\frac{p_0(f)}{Z_t^B}\exp\bigg(\sum_{s=1}^{t-1}\big[2\eta\epsilon_s(f(x_s, a_s)-f_*(x_s, a_s))-\eta\LS_s(f)\big]+\lambda\FG_t(f)\bigg),\label{eq:def_ptB}
\end{align}
where
\begin{align}
Z_t^B=\EE_{f\sim p_0}\exp\bigg(\sum_{s=1}^{t-1}\big[2\eta\epsilon_s(f(x_s,a_s)-f_*(x_s,a_s))-\eta\LS_s(f)\big]+\lambda\FG_t(f)\bigg).\label{eq:def_ZtB}
\end{align}
By the greedy action rule, the regret can also be written as
\begin{align*}
\sum_{t=1}^T[f_*(x_t, a_t^*)-f_*(x_t, a_t)]=\sum_{t=1}^T[f_t(x_t, a_t)-f_*(x_t, a_t)-\FG_t(f_t)].
\end{align*}

\subsection{Proof of Theorem~\ref{thm:A}}

\begin{proof}[Proof of Theorem~\ref{thm:A}]
In order to use the decoupling coefficient which is related to the expectation with respect to the joint distribution, we firstly perform the following decomposition of the regret:
\begin{align*}
\Regret(T)&=\underbrace{\sum_{t=1}^T\big\{(f_*(x_t, a_t^*)-f_*(x_t, a_t))-\EE[f_*(x_t, a_t^*)-f_*(x_t, a_t)|\cF_{t-1}]\big\}}_{I_1}\\
&\quad+\underbrace{\sum_{t=1}^T\EE[f_*(x_t, a_t^*)-f_*(x_t, a_t)|\cF_{t-1}]}_{I_2}.
\end{align*}
By Azuma-Hoeffding's inequality, we have with probability at least $1-\delta$,
\begin{align*}
I_1\le\sqrt{32T\log(1/\delta)}.
\end{align*}
For the term $I_2$, apply Definition~\ref{def:dc_A} conditionally on $\cF_{t-1},x_t,\cA_t$ and then average over the current context and action set. The auxiliary draw $\tilde f\sim p_t^A$ is independent of $a_t$ given this information. This gives
\begin{align*}
&\EE[f_*(x_t, a_t^*)-f_*(x_t, a_t)|\cF_{t-1}]\\
&=\EE[f_t(x_t, a_t)-f_*(x_t, a_t)-\FG_t(f_t)|\cF_{t-1}]\\
&\le\frac{\dc}{4\gamma}+\EE\Big[\EE_{\tilde f\sim p_t^A}[\gamma\LS_t(\tilde f)-\FG_t(\tilde f)]\Big|\cF_{t-1}\Big],
\end{align*}
so with probability at least $1-\delta$,
\begin{align*}
\Regret(T)\le\sqrt{32T\log(1/\delta)}+\frac{T\cdot\dc}{4\gamma}+\sum_{t=1}^T\EE\Big[\EE_{\tilde f\sim p_t^A}[\gamma\LS_t(\tilde f)-\FG_t(\tilde f)]\Big|\cF_{t-1}\Big].
\end{align*}
By Markov's inequality, for any $L_\delta$, we have
\begin{align*}
&\PP\bigg\{\lambda\sum_{t=1}^T\EE\Big[\EE_{\tilde f\sim p_t^A}[\gamma\LS_t(\tilde f)-\FG_t(\tilde f)]\Big|\cF_{t-1}\Big]\ge-\log Z_{T+1}^A+L_\delta\bigg\}\\
&=\PP\bigg\{Z_{T+1}^A\cdot\exp\bigg(\lambda\sum_{t=1}^T\EE\Big[\EE_{\tilde f\sim p_t^A}[\gamma\LS_t(\tilde f)-\FG_t(\tilde f)]\Big|\cF_{t-1}\Big]\bigg)\ge e^{L_\delta}\bigg\}\\
&\le e^{-L_\delta}\EE\bigg[Z_{T+1}^A\cdot\exp\bigg(\lambda\sum_{t=1}^T\EE\Big[\EE_{\tilde f\sim p_t^A}[\gamma\LS_t(\tilde f)-\FG_t(\tilde f)]\Big|\cF_{t-1}\Big]\bigg)\bigg]\\
&=e^{-L_\delta}\EE\bigg[\EE[Z_{T+1}^A|\cF_{T-1}]\cdot\exp\bigg(\lambda\sum_{t=1}^T\EE\Big[\EE_{\tilde f\sim p_t^A}[\gamma\LS_t(\tilde f)-\FG_t(\tilde f)]\Big|\cF_{t-1}\Big]\bigg)\bigg]
\end{align*}
For the term $\EE[Z_{T+1}^A|\cF_{T-1}]$, we use the following lemma:
\begin{lemma}\label{lemma:telescope}
Suppose that $0<\eta\le0.2$ and $\lambda>0$. Under Assumption~\ref{ass:subgaussian}, the Type-A normalizer satisfies
\begin{align*}
\EE[Z_{t+1}^A|\cF_{t-1}]
\le Z_t^A\cdot\exp\Big(\EE\Big[\EE_{f\sim p_t^A}[-0.25\eta\LS_t(f)+\lambda\FG_t(f)]\Big|\cF_{t-1}\Big]+4\lambda^2\Big).
\end{align*}
\end{lemma}
With Lemma~\ref{lemma:telescope}, we have
\begin{align*}
&\PP\bigg\{\lambda\sum_{t=1}^T\EE\Big[\EE_{\tilde f\sim p_t^A}[\gamma\LS_t(\tilde f)-\FG_t(\tilde f)]\Big|\cF_{t-1}\Big]\ge-\log Z_{T+1}^A+L_\delta\bigg\}\\
&\le e^{-L_\delta}\EE\bigg[Z_T^A\cdot\exp\bigg(\lambda\sum_{t=1}^{T-1}\EE\Big[\EE_{\tilde f\sim p_t^A}[\gamma\LS_t(\tilde f)-\FG_t(\tilde f)]\Big|\cF_{t-1}\Big]\bigg)\\
&\quad\cdot\exp\Big(\EE\Big[\EE_{\tilde f\sim p_T^A}[(\gamma\lambda-0.25\eta)\LS_T(\tilde f)]\Big|\cF_{T-1}\Big]+4\lambda^2\Big)\bigg]\\
&\le\cdots\\
&\le\exp(4T\lambda^2-L_\delta),
\end{align*}
where the second inequality holds because $\gamma\lambda\le0.25\eta$, and the last inequality follows by repeating this step down to $Z_1^A=1$. Therefore, taking
\begin{align*}
L_\delta=4T\lambda^2+\log(1/\delta),
\end{align*}
then the following event holds with probability at least $1-\delta$:
\begin{align*}
\sum_{t=1}^T\EE\Big[\EE_{\tilde f\sim p_t^A}[\gamma\LS_t(\tilde f)-\FG_t(\tilde f)]\Big|\cF_{t-1}\Big]\le\frac{-\log Z_{T+1}^A+\log(1/\delta)}{\lambda}+4T\lambda.
\end{align*}
The following lemma provides the bound for $-\log Z_{T+1}^A$:
\begin{lemma}\label{lemma:ZTA}
Under Assumption~\ref{ass:subgaussian}, for any $h>0$ with $p_0(\cF_h)>0$ and $\delta\in(0,1)$, with probability at least $1-\delta$, we have
\begin{align*}
-\log Z_{T+1}^A\le\eta h\sqrt{2T\log(1/\delta)}+\eta Th^2+\lambda Th+\log p_0^{-1}(\cF_h).
\end{align*}
\end{lemma}
With Lemma~\ref{lemma:ZTA}, combining all the above, with probability at least $1-3\delta$, we have
\begin{align*}
\Regret(T)
&\le(4+\eta h/\lambda)\sqrt{2T\log(1/\delta)}+\frac{T\cdot\dc}{4\gamma}+4T\lambda\\
&\quad+\frac{\eta Th^2}{\lambda}+Th+\frac{\log p_0^{-1}(\cF_h)+\log(1/\delta)}{\lambda}.
\end{align*}
Set
\begin{align*}
h=T^{-1},\quad\eta=0.2,\quad\lambda=\sqrt{\frac{\log p_0^{-1}(\cF_h)}{5T\cdot\dc}},\quad\gamma=0.05\lambda^{-1},
\end{align*}
then
\begin{align*}
\Regret(T)
&\le2\sqrt{5T\cdot\dc\cdot\log p_0^{-1}(\cF_h)}
+4\sqrt{\frac{T\log p_0^{-1}(\cF_h)}{5\dc}}\\
&\quad+\bigg(4\sqrt T+\sqrt{\frac{\dc}{5\log p_0^{-1}(\cF_h)}}\bigg)\sqrt{2\log(1/\delta)}\\
&\quad+\sqrt{\frac{5T\dc}{\log p_0^{-1}(\cF_h)}}\log(1/\delta)
+\sqrt{\frac{\dc}{5T\log p_0^{-1}(\cF_h)}}+1.
\end{align*}
\end{proof}

\subsection{Proof of Theorem~\ref{thm:B}}

\begin{proof}[Proof of Theorem~\ref{thm:B}]
According to the definition of the decoupling coefficient, for any $0<\gamma\le1$, the regret further satisfies
\begin{align*}
\Regret(T)\le\sum_{t=1}^T\bigg[\gamma\sum_{s=1}^{t-1}\LS_s(f_t)-\FG_t(f_t)\bigg]+\frac{\dc_T(\cF)}{4\gamma}.
\end{align*}
It then suffices to provide a high-probability upper bound of $\gamma\sum_{s=1}^{t-1}\LS_s(f_t)-\FG_t(f_t)$. By Markov's inequality, we have for any $C_t$,
\begin{align}
&\PP\bigg[\gamma\sum_{s=1}^{t-1}\LS_s(f_t)-\FG_t(f_t)\ge C_t\bigg]\nonumber\\
&=\PP\bigg[\exp\bigg(\gamma\lambda\sum_{s=1}^{t-1}\LS_s(f_t)-\lambda\FG_t(f_t)\bigg)\ge\exp(\lambda C_t)\bigg]\nonumber\\
&\le\exp(-\lambda C_t)\cdot\EE\exp\bigg(\gamma\lambda\sum_{s=1}^{t-1}\LS_s(f_t)-\lambda\FG_t(f_t)\bigg)\nonumber\\
&=\exp(-\lambda C_t)\cdot\EE\bigg[\frac1{Z_t^B}\EE_{\tilde f\sim p_0}\exp\bigg(\sum_{s=1}^{t-1}\big[2\eta\epsilon_s(\tilde f(x_s, a_s)-f_*(x_s, a_s))+(\gamma\lambda-\eta)\LS_s(\tilde f)\big]\bigg)\bigg].\label{eq:B_0}
\end{align}
where the last equality first conditions on $\cF_{t-1},x_t,\cA_t$ and then substitutes the posterior density in \eqref{eq:def_ptB}.
A key challenge is that the past noises $\epsilon_1,\ldots,\epsilon_{t-1}$ appear in both the numerator and the denominator $Z_t^B$. To solve this problem, we use the following lemma that provides a lower bound of $Z_t^B$:
\begin{lemma}\label{lemma:B_Zt}
For any $h>0$ with $p_0(\cF_h)>0$, the Type-B normalizer satisfies
\begin{align*}
\frac1{Z_t^B}\le\frac1{p_0(\cF_h)}\exp\bigg(\EE_{f\sim p_0^h}\bigg[\sum_{s=1}^{t-1}\big[-2\eta\epsilon_s(f(x_s, a_s)-f_*(x_s, a_s))+\eta h^2\big]+\lambda h\bigg]\bigg).
\end{align*}
\end{lemma}
Lemma~\ref{lemma:B_Zt} converts $Z_t^B$, which is the expectation of the exponential, into the exponential of the expectation, thus enabling the joint analysis of the past noises $\epsilon_1,\ldots,\epsilon_{t-1}$ in the numerator and the denominator. We thus have
\begin{align}
&\EE\bigg[\frac1{Z_t^B}\EE_{\tilde f\sim p_0}\exp\bigg(\sum_{s=1}^{t-1}\big[2\eta\epsilon_s(\tilde f(x_s, a_s)-f_*(x_s, a_s))+(\gamma\lambda-\eta)\LS_s(\tilde f)\big]\bigg)\bigg]\nonumber\\
&\le\frac{\exp(\lambda h)}{p_0(\cF_h)}\EE_{\tilde f\sim p_0}\EE\exp\bigg(\sum_{s=1}^{t-1}\big[2\eta\epsilon_s\{\tilde f(x_s, a_s)-\EE_{f\sim p_0^h}[f(x_s, a_s)]\}+(\gamma\lambda-\eta)\LS_s(\tilde f)+\eta h^2\big]\bigg).\label{eq:B_1}
\end{align}
We use the following lemma to further bound the right-hand side:
\begin{lemma}\label{lemma:B_supermartingle}
Under Assumption~\ref{ass:subgaussian}, suppose that $0<\eta\le0.5$, $\gamma,\lambda>0$, and $\gamma\lambda\le\eta-\eta^2$. For any fixed $\tilde f\in\cF$ and $h>0$ with $p_0(\cF_h)>0$, we have
\begin{align*}
&\EE\exp\bigg(\sum_{s=1}^{t-1}\big[2\eta\epsilon_s\{\tilde f(x_s, a_s)-\EE_{f\sim p_0^h}[f(x_s, a_s)]\}+(\gamma\lambda-\eta)\LS_s(\tilde f)+\eta h^2\big]\bigg)\\
&\le\exp(0.75(t-1)h^2).
\end{align*}
\end{lemma}
Plugging Lemma~\ref{lemma:B_supermartingle} and \eqref{eq:B_1} into \eqref{eq:B_0}, we have
\begin{align*}
\PP\bigg[\gamma\sum_{s=1}^{t-1}\LS_s(f_t)-\FG_t(f_t)\ge C_t\bigg]\le\frac{\exp(\lambda h+0.75(t-1)h^2-\lambda C_t)}{p_0(\cF_h)}.
\end{align*}
Let
\begin{align*}
C_t=h+\frac{0.75(t-1)h^2}{\lambda}+\frac{\log(p_0^{-1}(\cF_h))}{\lambda}+\frac{\log(T/\delta)}{\lambda},
\end{align*}
and $\cE_t$ be the event such that
\begin{align*}
\gamma\sum_{s=1}^{t-1}\LS_s(f_t)-\FG_t(f_t)\le C_t.
\end{align*}
Then $\cE_t$ holds with probability at least $1-\delta/T$. Taking the union bound, we have the good event $\cE:=\cap_{t\in[T]}\cE_t$ holds with probability at least $1-\delta$. Under the good event $\cE$, we have
\begin{align*}
\Regret(T)&\le\frac{\dc_T}{4\gamma}+\sum_{t=1}^TC_t\\
&=\frac{\dc_T}{4\gamma}+Th+\frac{3T(T-1)h^2}{8\lambda}+\frac{T\log(p_0^{-1}(\cF_h))}{\lambda}+\frac{T\log(T/\delta)}{\lambda}.
\end{align*}
Let
\begin{align*}
h=T^{-1},\quad\eta=0.5,\quad\lambda=\max\Big\{\frac14, \sqrt{T\dc_T^{-1}\log(p_0^{-1}(\cF_{1/T}))}\Big\},\quad\gamma=1/(4\lambda).
\end{align*}
The choice of $\lambda$ ensures that $0<\gamma\le1$ and $\gamma\lambda=\eta-\eta^2$. Since $\lambda\ge1/4$, the preceding bound gives
\begin{align*}
\Regret(T)\le\dc_T\lambda+3+\frac{T}{\lambda}\big[\log p_0^{-1}(\cF_{1/T})+\log(T/\delta)\big].
\end{align*}
To verify the final substitution, write $D=\log p_0^{-1}(\cF_{1/T})$ and $r=\sqrt{TD/\dc_T}$ within this proof. If $r\ge1/4$, substitute $\lambda=r$. If $r<1/4$ and $T\ge2$, Definition~\ref{def:dc_B} gives $\dc_T\le8T$, since $f_t(z_t)-f_*(z_t)\le2$. Consequently, the function
\begin{align*}
g(u)=\dc_Tu+\frac{T[D+\log(T/\delta)]}{u}
\end{align*}
satisfies, for $r\le u\le1/4$,
\begin{align*}
g'(u)=\dc_T-\frac{T[D+\log(T/\delta)]}{u^2}
\le8T-16T\log2<0.
\end{align*}
Thus $g(\lambda)\le g(r)$ in both cases, and
\begin{align*}
\Regret(T)\le2\sqrt{T\dc_T\log(p_0^{-1}(\cF_{1/T}))}+3+\sqrt{\frac{T\dc_T}{\log(p_0^{-1}(\cF_{1/T}))}}\log(T/\delta).
\end{align*}
For $T=1$, the same conclusion follows from $\Regret(1)\le2$.
\end{proof}

\section{Proof of Lemmas in Appendix~\ref{app:proof_main}}

\subsection{Proof of Lemma~\ref{lemma:telescope}}

\begin{proof}[Proof of Lemma~\ref{lemma:telescope}]
We first notice that
\begin{align*}
Z_{t+1}^A=Z_t^A\cdot\EE_{f\sim p_t^A}\exp(2\eta\epsilon_t(f(x_t, a_t)-f_*(x_t, a_t))-\eta\LS_t(f)+\lambda\FG_t(f)).
\end{align*}
Note that $Z_t^A$ and $p_t^A$ are $\cF_{t-1}$-measurable. Taking the conditional expectation, we have
\begin{align*}
\EE[Z_{t+1}^A|\cF_{t-1}]&=Z_t^A\cdot\EE\Big[\EE_{f\sim p_t^A}\EE\big[\exp(2\eta\epsilon_t(f(x_t, a_t)-f_*(x_t, a_t))-\eta\LS_t(f)+\lambda\FG_t(f))\big|\cG_t\big]\Big|\cF_{t-1}\Big]\\
&\le Z_t^A\cdot\EE\Big[\EE_{f\sim p_t^A}\exp((\eta^2/2-\eta)\LS_t(f)+\lambda\FG_t(f))\Big|\cF_{t-1}\Big]\\
&\le Z_t^A\cdot\EE\Big[\EE_{f\sim p_t^A}\exp(-0.9\eta\LS_t(f)+\lambda\FG_t(f))\Big|\cF_{t-1}\Big],
\end{align*}
where the first inequality holds due to Assumption~\ref{ass:subgaussian}, and the second inequality holds because $\eta\le0.2$. To bound this exponential moment in terms of the mean loss and mean feel-good term, we use the Cauchy-Schwarz inequality:
\begin{align}
&\EE\Big[\EE_{f\sim p_t^A}\exp(-0.9\eta\LS_t(f)+\lambda\FG_t(f))\Big|\cF_{t-1}\Big]\nonumber\\
&\le\sqrt{\EE\Big[\EE_{f\sim p_t^A}\exp(-1.8\eta\LS_t(f))\Big|\cF_{t-1}\Big]}\cdot\sqrt{\EE\Big[\EE_{f\sim p_t^A}\exp(2\lambda\FG_t(f))\Big|\cF_{t-1}\Big]}.\label{eq:increment}
\end{align}
Since $e^z\le1+z+z^2/2$ for $z\le0$, we have
\begin{align}
&\EE\Big[\EE_{f\sim p_t^A}\exp(-1.8\eta\LS_t(f))\Big|\cF_{t-1}\Big]\nonumber\\
&\le\EE\Big[\EE_{f\sim p_t^A}[1-1.8\eta\LS_t(f)+1.62\eta^2\LS_t^2(f)]\Big|\cF_{t-1}\Big]\nonumber\\
&\le\EE\Big[\EE_{f\sim p_t^A}[1-1.8\eta\LS_t(f)+1.296\eta\LS_t(f)]\Big|\cF_{t-1}\Big]\nonumber\\
&\le1-0.5\eta\EE\Big[\EE_{f\sim p_t^A}[\LS_t(f)]\Big|\cF_{t-1}\Big]\nonumber\\
&\le\exp\Big(-0.5\eta\EE\Big[\EE_{f\sim p_t^A}[\LS_t(f)]\Big|\cF_{t-1}\Big]\Big),\label{eq:LS}
\end{align}
where the second inequality holds because $\eta\le0.2$ and $\LS_t(f)\le4$, and the last inequality holds because $1+z\le e^z$. For the term related to $\FG_t(f)$, by Hoeffding's lemma, since $\FG_t(f)\in[-2, 2]$, we have
\begin{align}
\EE\Big[\EE_{f\sim p_t^A}\exp(2\lambda\FG_t(f))\Big|\cF_{t-1}\Big]\le\exp(2\lambda\EE\Big[\EE_{f\sim p_t^A}[\FG_t(f)]\Big|\cF_{t-1}\Big]+8\lambda^2\Big).\label{eq:FG}
\end{align}
Plugging \eqref{eq:LS} and \eqref{eq:FG} into \eqref{eq:increment}, we have
\begin{align*}
&\EE\Big[\EE_{f\sim p_t^A}\exp(-0.9\eta\LS_t(f)+\lambda\FG_t(f))\Big|\cF_{t-1}\Big]\\
&\le\exp\Big(\EE\Big[\EE_{f\sim p_t^A}[-0.25\eta\LS_t(f)+\lambda\FG_t(f)]\Big|\cF_{t-1}\Big]+4\lambda^2\Big).
\end{align*}
\end{proof}

\subsection{Proof of Lemma~\ref{lemma:ZTA}}

\begin{proof}[Proof of Lemma~\ref{lemma:ZTA}]
We use Lemma~\ref{lemma:KL_reg} with
\begin{align*}
P=p_0^h,\quad P_0=p_0,\quad U(f)=\sum_{t=1}^T\big[-2\eta\epsilon_t(f(x_t, a_t)-f_*(x_t, a_t))+\eta\LS_t(f)-\lambda\FG_t(f)\big].
\end{align*}
Then Lemma~\ref{lemma:KL_reg} gives
\begin{align*}
-\log Z_{T+1}^A\le\EE_{f\sim p_0^h}\bigg[\sum_{t=1}^T\big[-2\eta\epsilon_t(f(x_t, a_t)-f_*(x_t, a_t))+\eta\LS_t(f)-\lambda\FG_t(f)\big]\bigg]+\KL(p_0^h\|p_0).
\end{align*}
Due to the definition of $\cF_h$, for any $f\in\cF_h$, we have
\begin{align*}
\LS_t(f)\le h^2,\quad-\FG_t(f)=f_*(x_t, a_t^*)-\max_{a\in\cA_t}f(x_t, a)\le f_*(x_t, a_t^*)-f(x_t, a_t^*)\le h.
\end{align*}
For the KL divergence term, we have
\begin{align*}
\KL(p_0^h\|p_0)=\EE_{f\sim p_0^h}\log\frac{p_0^h(f)}{p_0(f)}=\log p_0^{-1}(\cF_h).
\end{align*}
Additionally, since $|\EE_{f\sim p_0^h}f(x_t, a_t)-f_*(x_t, a_t)|\le h$, Lemma~\ref{lemma:subgaussian} gives that with probability at least $1-\delta$, we have
\begin{align*}
\sum_{t=1}^T\epsilon_t\cdot\EE_{f\sim p_0^h}[f_*(x_t, a_t)-f(x_t, a_t)]\le h\sqrt{T/2\log(1/\delta)}.
\end{align*}
Combining all the above, we have
\begin{align*}
-\log Z_{T+1}^A\le\eta h\sqrt{2T\log(1/\delta)}+\eta Th^2+\lambda Th+\log p_0^{-1}(\cF_h).
\end{align*}
\end{proof}

\subsection{Proof of Lemma~\ref{lemma:B_Zt}}

\begin{proof}[Proof of Lemma~\ref{lemma:B_Zt}]
We use Lemma~\ref{lemma:KL_reg} with
\begin{align*}
P=p_0^h,\quad P_0=p_0,\quad U(f)=\sum_{s=1}^{t-1}\big[-2\eta\epsilon_s(f(x_s, a_s)-f_*(x_s, a_s))+\eta\LS_s(f)\big]-\lambda\FG_t(f).
\end{align*}
Then Lemma~\ref{lemma:KL_reg} gives
\begin{align*}
\log\frac1{Z_t^B}\le\EE_{f\sim p_0^h}\bigg[\sum_{s=1}^{t-1}\big[-2\eta\epsilon_s(f(x_s, a_s)-f_*(x_s, a_s))+\eta\LS_s(f)\big]-\lambda\FG_t(f)\bigg]+\KL(p_0^h\|p_0).
\end{align*}
Due to the definition of $\cF_h$, for any $f\in\cF_h$, we have
\begin{align*}
\LS_s(f)\le h^2,\quad-\FG_t(f)=f_*(x_t, a_t^*)-\max_{a\in\cA_t}f(x_t, a)\le f_*(x_t, a_t^*)-f(x_t, a_t^*)\le h.
\end{align*}
For the KL divergence term, we have
\begin{align*}
\KL(p_0^h\|p_0)=\EE_{f\sim p_0^h}\log\frac{p_0^h(f)}{p_0(f)}=\log p_0^{-1}(\cF_h).
\end{align*}
Combining all the above, we have
\begin{align*}
\frac1{Z_t^B}\le\frac1{p_0(\cF_h)}\exp\bigg(\EE_{f\sim p_0^h}\bigg[\sum_{s=1}^{t-1}\big[-2\eta\epsilon_s(f(x_s, a_s)-f_*(x_s, a_s))+\eta h^2\big]+\lambda h\bigg]\bigg).
\end{align*}
\end{proof}

\subsection{Proof of Lemma~\ref{lemma:B_supermartingle}}

\begin{proof}[Proof of Lemma~\ref{lemma:B_supermartingle}]
We denote
\begin{align*}
w_s=2\eta\epsilon_s\{\tilde f(x_s, a_s)-\EE_{f\sim p_0^h}[f(x_s, a_s)]\}+(\gamma\lambda-\eta)\LS_s(\tilde f)+\eta h^2,
\end{align*}
Then $w_s$ is $\cF_s$-measurable. For $1\le k<t$, we have
\begin{align*}
\EE\exp\bigg(\sum_{s=1}^{k}w_s\bigg)&=\EE\bigg\{\exp\bigg(\sum_{s=1}^{k-1}w_s\bigg)\cdot\EE[\exp(w_k)|\cF_{k-1}]\bigg\},
\end{align*}
where
\begin{align*}
&\EE[\exp(w_k)|\cF_{k-1}]\\
&=\EE\{\EE[\exp(w_k)|\cG_k]|\cF_{k-1}\}\\
&\le\EE\Big[\exp\Big(\frac{\eta^2}{2}\{\tilde f(x_k, a_k)-\EE_{f\sim p_0^h}[f(x_k, a_k)]\}^2+(\gamma\lambda-\eta)\LS_k(\tilde f)+\eta h^2\Big)\Big|\cF_{k-1}\Big]\\
&\le\EE\Big[\exp\Big(\eta^2\{\EE_{f\sim p_0^h}[f_*(x_k, a_k)-f(x_k, a_k)]\}^2+(\gamma\lambda-\eta+\eta^2)\LS_k(\tilde f)+\eta h^2\Big)\Big|\cF_{k-1}\Big]\\
&\le\exp(0.75h^2),
\end{align*}
where the first inequality holds due to Assumption~\ref{ass:subgaussian}, the second inequality holds from $(u+v)^2\le2u^2+2v^2$, and the last inequality holds because $|\EE_{f\sim p_0^h}[f_*(x_k,a_k)-f(x_k,a_k)]|\le h$, $\eta+\eta^2\le0.75$, and $\gamma\lambda\le\eta-\eta^2$. Iteratively using the argument above, we have
\begin{align*}
\EE\exp\bigg(\sum_{s=1}^{t-1}w_s\bigg)\le\exp(0.75(t-1)h^2).
\end{align*}
\end{proof}

\section{Auxiliary Lemmas}

\begin{lemma}[Donsker--Varadhan duality, see e.g., Proposition 7.16 in \citet{zhang2023mathematical}]\label{lemma:KL_reg}
Let $(\cX, \cF, P_0)$ be a probability space and $U(x)$ be a bounded measurable function. Then for any distribution $P$ on $(\cX, \cF)$, we have
\begin{align*}
\EE_{x\sim P}[U(x)]+\KL(P\|P_0)\ge-\log\EE_{x\sim P_0}\exp(-U(x)),
\end{align*}
and equality holds for the distribution with density $\exp(-U)/\EE_{P_0}\exp(-U)$ with respect to $P_0$.
\end{lemma}
\begin{lemma}\label{lemma:subgaussian}
Let $\{\cF_t\}$ and $\{\cG_t\}$ be filtrations such that $\cF_{t-1}\subseteq\cG_t\subseteq\cF_t$. Let $\epsilon_t$ be $\cF_t$-measurable and satisfy
\begin{align*}
\log\EE[\exp(\lambda\epsilon_t)|\cG_t]\le\lambda^2/8
\qquad\text{for every $\lambda\in\RR$}.
\end{align*}
Let $\mu_t$ be $\cG_t$-measurable with $|\mu_t|\le C$ almost surely, where $C>0$. For every $M\ge0$, we have
\begin{align*}
\PP\bigg\{\sum_{t=1}^T\mu_t\epsilon_t\ge M\bigg\}\le\exp\Big(-\frac{2M^2}{C^2T}\Big).
\end{align*}
Therefore, for any $\delta\in(0,1)$, with probability at least $1-\delta$, we have
\begin{align*}
\sum_{t=1}^T\mu_t\epsilon_t\le C\sqrt{\frac T2\log(1/\delta)}.
\end{align*}
\end{lemma}
\begin{proof}[Proof of Lemma~\ref{lemma:subgaussian}]
For $M>0$ and $\gamma>0$, Markov's inequality gives
\begin{align*}
\PP\bigg\{\sum_{t=1}^T\mu_t\epsilon_t\ge M\bigg\}&=\PP\bigg\{\exp\bigg(\gamma\sum_{t=1}^T\mu_t\epsilon_t\bigg)\ge\exp(\gamma M)\bigg\}\\
&\le\exp(-\gamma M)\EE\exp\bigg(\gamma\sum_{t=1}^T\mu_t\epsilon_t\bigg)\\
&=\exp(-\gamma M)\EE\bigg\{\EE\bigg[\exp\bigg(\gamma\sum_{t=1}^T\mu_t\epsilon_t\bigg)\bigg|\cG_T\bigg]\bigg\}\\
&\le\exp(-\gamma M)\EE\exp\bigg(\gamma\sum_{t=1}^{T-1}\mu_t\epsilon_t+\frac{\gamma^2\mu_T^2}8\bigg)\\
&\le\exp(-\gamma M)\EE\exp\bigg(\gamma\sum_{t=1}^{T-1}\mu_t\epsilon_t+\frac{\gamma^2C^2}8\bigg)\\
&\le\exp(-\gamma M+\gamma^2C^2T/8).
\end{align*}
where the second inequality holds by the conditional moment bound given $\cG_T$, the third inequality holds because $|\mu_t|\le C$ almost surely, and the last inequality holds by recursively applying the inequalities. Taking $\gamma=4M/(C^2T)$ gives
\begin{align*}
\PP\bigg\{\sum_{t=1}^T\mu_t\epsilon_t\ge M\bigg\}\le\exp\Big(-\frac{2M^2}{C^2T}\Big).
\end{align*}
For $M=0$, the asserted bound is $1$.
\end{proof}

\bibliography{arxiv}
\bibliographystyle{ims}

\end{document}